\documentclass[conference]{IEEEtran}
\usepackage[T1]{fontenc}
\usepackage{amsmath,amssymb}
\usepackage{graphicx}
\usepackage{booktabs}
\usepackage{multirow}
\usepackage{xcolor}
\usepackage{cite}
\usepackage[hidelinks]{hyperref}
\graphicspath{{figures/}}

\newcommand{\base}{SuperMimic}
\newcommand{\med}{\operatorname{med}}
\newcommand{\E}{\mathbb{E}}
\newcommand{\pp}[2]{\frac{\partial #1}{\partial #2}}

\begin{document}

\title{Trust What Repeats: Capped World-Model\\Policy Gradients for Humanoid Motion Tracking}

\author{\IEEEauthorblockN{Xuzhe Dang and Stefan Edelkamp}
\IEEEauthorblockA{Faculty of Electrical Engineering, Czech Technical University in Prague\\
Faculty of Mathematics and Physics, Charles University\\
Prague, Czech Republic\\
dangxuzh@fel.cvut.cz, stefan.edelkamp@gmail.com}}

\maketitle

\begin{abstract}
Training a humanoid motion tracker by backpropagating imagined returns through a learned world model is fast, because
the model gives a first-order policy gradient. We ask how much of that gradient holds in the simulator. On a SuperMimic
tracker for the Unitree G1, the $0.5\%$ of imagined rollouts with the largest gradients carry $22\%$ of the summed gradient
norm, yet neither a second rollout from the same state nor the simulator confirms them. When the actions are moved
along the model's gradient, the simulator returns $96\%$ of the predicted change in return where the gradient norm is
below the batch median, $45\%$ between $3$ and $10$ times the median, and $1\%$ above $30$ times; the part of a
gradient that either check confirms stops growing at a few times the median. We therefore cap each rollout's
action-gradient norm at $10$ times the batch median, draw imagination starts in proportion to their predicted gradient
norm so that the cap removes less, and reset episodes where the world model and critic predict the lowest return.
Trained on 223 hours of motion and tested on 353 motions that no actor performed in training, the method raises the
normalized return of SuperMimic, the baseline it builds on, from $0.948$ to $0.960$ and the lowest $1\%$ of test clips
from $0.64$ to $0.75$, cuts the test clips that fall persistently from $14$--$19$ to $3$--$7$, and reaches the
baseline's final return in $44\%$ of the baseline's iterations and half its training time. On LAFAN1, trained and
evaluated on the same clips as prior work, it reaches $0.872$ against $0.840$. Code: \url{https://github.com/XZDang13/trustmimic}.

\end{abstract}

\section{Introduction}

A general motion tracker turns a library of human motion into a single policy that a humanoid can run: given the
next frames of a reference, the policy outputs joint targets that make the robot follow it. Such trackers are the
low-level layer under teleoperation, motion generation and learned skills, and their quality is set by how many
motions they follow without falling and by how long they take to train. With model-free reinforcement learning the
second number is large: trackers for libraries of a hundred hours and more are trained for days on many
GPUs~\cite{luo2023phc,liao2025beyondmimic,nvidia2025sonic}.

\base{}~\cite{liao2026supermimic} changed that. It fits a world model to the simulator's transitions and improves the
policy by backpropagating the imagined return through the model, a first-order gradient that needs no likelihood
ratio and no exploration in action space. The authors report the return of a strong model-free tracker after hours
on one GPU. The price of a first-order gradient is known from differentiable simulation: through contact and over
long horizons the gradient of a rollout can be large, wrong, or both~\cite{parmas2018pipps,metz2021gradients,suh2022differentiable}, and a model that is
accurate in its predictions need not be accurate in its derivatives.

This paper measures that price for motion tracking and removes most of it. It rests on two measurements on a
trained tracker.

\emph{A few rollouts carry the gradient.} Among the imagined rollouts of a trained SuperMimic tracker, the $0.5\%$
whose action-gradient norm $G$ exceeds $30$ times the batch median hold $22\%$ of the summed norm and $96\%$ of the
summed squared norm, while $86\%$ of rollouts lie below $3$ times the median. Heavy tails of this kind are known for gradients through
simulators~\cite{suh2022differentiable} and were reported for humanoid tracking in concurrent
work~\cite{lu2026nablamimic}; what has not been asked is whether the tail is signal.

\emph{The large gradients are not backed.} We roll the same start state out twice under independent sensor noise
and ask how much of one rollout's gradient the other reproduces; and we set the simulator to states from the replay
buffer, step it along the model's gradient, and ask how much of the predicted change of the return appears. The two
tests agree (Fig.~\ref{fig:worth}). Up to about $3$ times the median, both confirm most of a gradient's norm. Beyond
that the confirmed part stops growing: from $5$ times on, both confirm about $2$ to $5$ times the median whatever
the norm, and the simulator returns $45\%$ of the
model's predicted change between $3$ and $10$ times the median, $7\%$ to $8\%$ between $10$ and $30$ times and at
most $1\%$ above, for the baseline's tracker and ours alike.

The part of a rollout's gradient that both tests confirm is thus roughly $\min(G, k \cdot \med G)$, with $\med G$
the batch median and $k$ a few units, and we make that the update rule: the norm of each rollout's action gradient
is capped at $k = 10$ times the batch median. The cap acts as a trust region on the model's derivative, set per
rollout and measured against the batch's own median. Two further changes go with it. Start states whose gradients
the cap would cut are drawn more often and weighted down, which leaves the expected gradient unchanged and gives
the cap less to remove. And episodes are reset where the world model and critic predict the lowest return, ranked
over the whole library every 2{,}000 iterations, which directs data to the motions the policy cannot yet follow and
finds them without first visiting them.

We evaluate on a split that no tracker of this kind has reported, to our knowledge: BONES-SEED retargeted to the
Unitree G1, divided by motion, so that no test motion was performed by any actor in training. On 11{,}828 test clips of 353 unseen motions and three seeds, the method raises the
normalized return from $0.948$ to $0.960$ and the lowest $1\%$ of clips from $0.64$ to $0.75$, and cuts the clips
that fall in at least half of the last six snapshots from $14$--$19$ to $3$--$7$ (Fig.~\ref{fig:rollouts}). It passes the baseline's final test return at about $11{,}000$ of $25{,}000$
iterations, after $1.2$ to $1.3$ hours against $2.5$ to $2.6$. The cap is what raises the mean and the model-ranked resets
what raise the tail; the start draw adds little at this scale. Doubling the baseline's learning rate gives it about a seventh of the
method's gain at $15{,}000$ iterations and little of the drop in failing clips. On LAFAN1, the motion set on which this line of work reports results, the method reaches $0.872$
against $0.840$ and passes the baseline's final return at $7{,}000$ to $7{,}500$ of $15{,}000$ iterations.

Our contributions are:
\begin{itemize}
\item two tests of a world model's policy gradient, one against a redraw of the same start and one against the
simulator, and the finding that both level off at a few times the batch median of the gradient norm;
\item an update built on that finding: a cap on each rollout's action-gradient norm, a start distribution drawn
by gradient norm, which reduces what the cap removes without adding bias but adds little once the resets are
model-ranked, and a reset distribution for motion tracking that is
ranked by the world model and critic instead of by visits;
\item an evaluation of world-model motion tracking on motions unseen in training, with ablations of each part, on
two versions of the Unitree G1, and a comparison on LAFAN1.
\end{itemize}
The code, with the scripts for training, evaluation and the train-test split, is available at
\url{https://github.com/XZDang13/trustmimic}.

\section{Related Work}
\label{sec:related}

\emph{Motion tracking at scale.} Since DeepMimic~\cite{peng2018deepmimic}, physics-based trackers are trained with
model-free reinforcement learning from random reference phases with early termination. Scaling them to large
libraries has relied on where episodes start. PHC~\cite{luo2023phc} evaluates the whole dataset and samples the
clips the policy fails; BeyondMimic~\cite{liao2025beyondmimic} and SONIC~\cite{nvidia2025sonic} keep a failure rate
per bin of a clip and sample bins by it; GMT~\cite{chen2025gmt} adapts sampling to tracking error.
\base{}~\cite{liao2026supermimic} keeps a return per bin from the simulator's rollouts. All of these score a bin
from episodes that visited it. We score every bin through the world model and critic without visiting it.

\emph{First-order policy gradients.} Differentiating a simulator or a learned model gives low-variance policy
gradients~\cite{hafner2020dreamer,xu2022shac}, and a known set of failures: the gradient's variance grows with the
horizon in chaotic systems~\cite{parmas2018pipps,metz2021gradients}, and through contact a first-order estimate can
be biased by rare large samples~\cite{suh2022differentiable}. The usual remedies are short horizons closed by a
critic and a clip on the norm of the batch gradient~\cite{xu2022shac,pascanu2013difficulty}, or a horizon adapted to
contact stiffness~\cite{georgiev2024ahac}. \base{}~\cite{liao2026supermimic} showed that a learned model makes the
approach practical for humanoid tracking. Our redraw and simulator tests address the question of
\cite{suh2022differentiable} directly for a learned model: whether the rare large samples carry the true gradient.
In our setting they do not.

\emph{Clipping and truncation.} The operator we use, a clip of each sample's gradient at a threshold set by a
quantile of the norms, is that of differentially private training~\cite{abadi2016dpsgd,andrew2021adaptive}, used
there for privacy and here for accuracy; what is ours is the unit it acts on, a rollout's action gradient, and the
evidence for its level. Clipping biases the update~\cite{koloskova2023clipping}, and the trade between that bias and
the variance removed is the one studied for truncated importance
weights~\cite{ionides2008truncated,su2020shrinkage}; \cite{su2020shrinkage} derive clipping as the minimizer of a
worst-case bias plus the second moment, an argument that carries over to gradients.

\emph{Sampling by gradient norm.} Drawing training examples in proportion to their gradient norm and weighting them
by the inverse is the variance-optimal importance sampling for stochastic gradient
descent~\cite{needell2014stochastic,zhao2015stochastic,alain2015variance,katharopoulos2018not}, and prioritized
replay~\cite{schaul2016per} does the same with the temporal-difference error. We apply it to the start states of
imagined rollouts, with the norm predicted from the critic's value, and note that its purpose here is different:
it leaves a cap less to remove.

\emph{Curricula from critics and models.} Prioritized level replay~\cite{jiang2021plr} ranks tasks by a value
loss from actual rollouts. ProCuRL~\cite{tzannetos2023procurl} and SPaCE~\cite{eimer2021space} score every task
with one forward pass of the critic, which is our ranking at a horizon of zero. Closest in mechanism are methods
that roll the policy through a world model to score goals or start states~\cite{hu2023peg,guzel2025imac}, and the
score itself is a model-based value expansion~\cite{feinberg2018mve}. We are not aware of a reset distribution for
motion tracking that is ranked by a learned model.

\emph{Weight averaging.} Evaluating an average of the weights that never acts in
training~\cite{nikishin2018swa} is standard in imitation learning and uncommon in reinforcement-learning trackers.
We use it for both methods and claim nothing for it.

\emph{Concurrent work.} $\nabla$Mimic~\cite{lu2026nablamimic} reports heavy-tailed first-order gradients for
humanoid tracking and clips the action gradient of each step and environment at a multiple of the batch median
inside the backward pass; we cap once per rollout and test what the capped part is worth.
InstantTrack~\cite{dao2026instanttrack} and BlinkMimic~\cite{chen2026blinkmimic} speed up \base{}'s line with a
different optimizer, a residual policy, GPU graph capture and, in the latter, resets from failures of the deployed
policy in simulation, and train $40$ LAFAN1 clips in minutes; InstantMimic~\cite{choi2026instantmimic} is a
GPU-native system for model-free tracking. These changes are largely orthogonal to ours. None of them evaluates on
motions held out from training.

\section{Background}
\label{sec:background}

\subsection{Motion tracking with a world model}

We build on \base{}~\cite{liao2026supermimic} and keep its environment, networks and losses, in our own
implementation. The robot is a Unitree G1, simulated in Newton~\cite{newton} with its MuJoCo
Warp solver~\cite{todorov2012mujoco,mujocowarp} at 200\,Hz and controlled at 50\,Hz through joint position targets. A motion library holds retargeted
clips; an episode starts at a frame of a clip and ends when the clip ends or when the robot's height or tilt leaves
a band around the reference. The reward combines exponential kernels on the tracking errors of body positions,
orientations and velocities with small action penalties. As in~\cite{liao2026supermimic} we report the \emph{normalized
return} of a clip: the sum of its rewards when played from the first frame to the last, over the largest value that
sum can take, so that $1$ is perfect and a fall forfeits the rest of the clip.

The policy $\pi_\theta$ sees a history of $10$ noisy proprioceptive observations and the next $10$ reference frames,
each of them the reference's joint positions and velocities and its root orientation relative to the robot, but not
the root's position, and outputs an action $a_t$. A world model $f_\omega$ predicts the next simulator state from the state and the
action, $s_{t+1} = f_\omega(s_t, a_t)$, and is fit to transitions that $4{,}096$ parallel environments write to a
replay buffer. A critic $V_\psi$ is fit to $\lambda$-returns of imagined rollouts. Reward, termination and the
observations are known functions of the state, so nothing but the dynamics is learned.

\subsection{The first-order policy gradient}

Each iteration draws $B$ states from the replay buffer and rolls each out for $H$ steps with the policy inside the
model. With $c_t \in [0,1]$ a smooth survival probability, a smoothed version of the termination test computed from the
model's state, the objective of rollout $b$ is
\begin{equation}
J_b(\theta) = \sum_{t=0}^{H-1} \gamma^t\, \bar c_{b,t}\, r_{b,t} \;+\; \gamma^{H} \bar c_{b,H}\, V_\psi(s_{b,H}),
\label{eq:objective}
\end{equation}
where $\bar c_{b,t} = \prod_{u<t} c_{b,u}$ is the probability of surviving to step $t$. A rollout that reaches
the end of its clip before step $H$ stops there and is closed by the critic in the same way. The policy ascends
$\nabla_\theta \frac{1}{B}\sum_b J_b$, computed by backpropagation through the $H$ model steps. \base{} uses
$H = 32$ and $B = 4{,}096$.

Write $g_{b,t} = \mathrm{d}J_b / \mathrm{d}a_{b,t}$ for the total derivative of a rollout's objective with respect
to its action at step $t$, including the change of later actions through the policy. The policy gradient is then a sum over
rollouts and steps of these action gradients pulled back through the policy at the observation $o_{b,t}$ it saw,
\begin{equation}
\nabla_\theta J = \frac{1}{B} \sum_{b} \sum_{t} \Big(\pp{\pi_\theta(o_{b,t})}{\theta}\Big)^{\!\top} g_{b,t},
\label{eq:action-gradient}
\end{equation}
and we call $G_b = \big(\sum_t \lVert g_{b,t} \rVert^2\big)^{1/2}$ the \emph{gradient norm of rollout $b$}. It is
the quantity this paper measures and caps.

\subsection{Where episodes start}

Episodes that end are reset to a frame of the library. \base{} divides clips into bins of about one second and keeps
for each bin a score from the returns of the episodes started there. A reset goes to a uniformly drawn bin with
probability $0.5$, to a bin drawn with a probability that grows as its score falls with probability $0.4$, and to a
bin drawn by how often bins were drawn so far with probability $0.1$. The score of a bin changes only when the bin is visited.

\section{How much of the model's gradient is real?}
\label{sec:analysis}

The measurements of this section are made on trained trackers with the policy and the models held fixed. We make
them on two policies trained on the full 29-dof library for $25{,}000$ iterations, the baseline's and the one the
method below trains, and they agree. Where one number is given it is of the method's policy, since an update rule
has to hold at the policies it produces, and the baseline's follows in brackets or beside it. The same conclusions
hold for policies trained on a subset of the 23-dof library under other update rules and at an earlier checkpoint
(Appendix~\ref{app:analysis}).

\subsection{A few rollouts carry the gradient}

Over $65{,}536$ rollouts of $H = 64$ steps, the horizon the method trains with, from states of the replay buffer, $93\%$ have a gradient norm $G_b$ below $3$ times
the median and $0.23\%$ above $30$ times. That last fifth of a percent holds $13\%$ of $\sum_b G_b$ and $91\%$ of
$\sum_b G_b^2$. The baseline's policy has the heavier tail: $86\%$ of its rollouts lie below $3$ times the median,
and the $0.53\%$ above $30$ times hold $22\%$ of $\sum_b G_b$ and $96\%$ of $\sum_b G_b^2$. A batch mean of the gradients is therefore set by a handful of rollouts, and its effective sample
size is a small fraction of $B$. The same has been observed for gradients through differentiable
simulators~\cite{parmas2018pipps,metz2021gradients,suh2022differentiable}; the question for a learned model is whether those few rollouts are right.

\subsection{Test 1: does the gradient repeat?}

An imagined rollout is a deterministic function of its start state except for the sensor noise added to the
policy's observations. We roll each of $32{,}768$ start states out twice under independent noise, with action
gradients $g$ and $g'$ of norms $G$ and $G'$. If $g = \mu(s) + n$ with $\mu(s)$ the part of the gradient that
belongs to the start and $n$ the part that the noise adds, then $\E\langle g, g'\rangle = \lVert\mu(s)\rVert^2$.
The scalar $\rho$ that makes $\rho\, g$ the best estimate of $\mu$ in mean square is
$\rho = \langle g, \mu\rangle / G^2$, and we call the norm $\rho\,G$ of that estimate the \emph{worth} of a rollout
of norm $G$:
\begin{equation}
c^\star(G) \;=\; \E\big[\, G' \cos\angle(g, g') \;\big|\; G \,\big],
\label{eq:worth}
\end{equation}
the expected projection of the redraw onto it. Fig.~\ref{fig:worth} plots $c^\star$ by band of $G$. Up to $3$ times
the median a rollout is worth $0.79$ to $0.92$ of its norm. From $5$ to $30$ times the median it is worth
$2.6$ medians ($95\%$ interval $1.6$ to $3.4$), while the mean norm in that band is $10$ medians, and $5.1$ ($4.0$ to $6.7$) for the
baseline's policy. Above $30$ times the worth is not measured well: these rollouts are $0.23\%$ of the sample, and
their estimates scatter from $-4$ to $7$ medians.

\subsection{Test 2: does the simulator agree?}

The first test asks whether the model's gradient is consistent with itself. The second asks the simulator. From
$5{,}456$ replay states we take the model's action gradient $g$ of the $64$-step rollout, set the simulator to the
same state, and play the policy with its actions at all $64$ steps shifted by $\pm\delta$, $\delta = \epsilon\, g / G$, where
$\epsilon$ is chosen so that the largest element of $\delta$ is $0.05$, half the exploration noise; a step five times
smaller gives the same trust in every band of Table~\ref{tab:trust} to within $0.06$, for both methods (Appendix~\ref{app:analysis}). Sensor noise is
off and the two perturbed runs share everything else. With $J$ the objective of Eq.~\eqref{eq:objective} evaluated on the
simulator's states, the model predicts a change of $2\epsilon G$ and we define
\begin{equation}
\text{trust} \;=\; \frac{J(+\delta) - J(-\delta)}{2\,\epsilon\, G}.
\label{eq:trust}
\end{equation}
A trust of $1$ means the simulator returns the whole predicted change. Three controls bound the test's own error:
a state written to the simulator reads back to $10^{-7}$; two identical copies differ by $0.1\%$ to $1.1\%$ of the
predicted change; and the unperturbed return of the model's rollout is within $1\%$ of the simulator's.

\begin{figure}[t]
\centering
\includegraphics[width=\columnwidth]{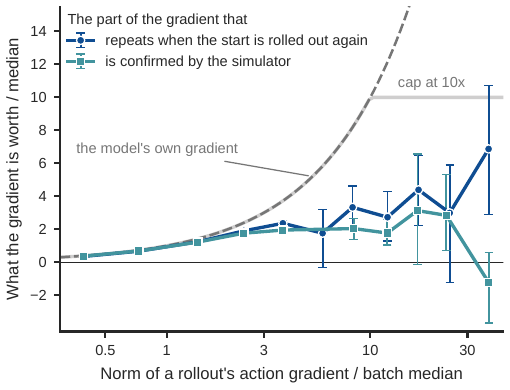}
\caption{What a rollout's action gradient is worth, against its norm, both in units of the batch median. Blue circles:
the expected projection of a second rollout from the same start state (Eq.~\ref{eq:worth}). Teal squares: the slope
the simulator confirms along the model's gradient, trust times $G$. Bars are $95\%$ intervals. Both follow the dashed
line, on which a gradient is worth its whole norm, up to about $3$ times the median and level off after; the grey
line is a cap at $10$ times the median.}
\label{fig:worth}
\end{figure}

Table~\ref{tab:trust} gives the result. Below the median the simulator returns $95\%$ of what the model predicts.
The share falls to $45\%$ between $3$ and $10$ times the median, $7\%$ between $10$ and $30$, and nothing above,
and the baseline's policy gives the same numbers to within $0.05$.
The direction survives longer than the size: above $10$ times the median the simulator's change still has the
predicted sign on $60\%$ to $72\%$ of states (third row). Two things are lost. Between $3$ and $10$ times the median the model's own finite difference still agrees with its
gradient ($0.87$), so the model is smooth there and simply wrong about the simulator. Above $10$ times the model's
finite difference departs from its gradient as well ($0.22$, then $0.03$): the gradient is a derivative at a point
that says little about a change of the actions as large as the perturbation. The slope the simulator confirms,
trust times $G$, is $2.3$ medians ($1.8$ to $2.9$) from $5$ to $30$ times the median, and $2.2$ ($1.7$ to $2.6$) for the
baseline's policy, flat in $G$ like the worth of
Eq.~\eqref{eq:worth} (Fig.~\ref{fig:worth}).

\begin{table}[t]
\centering
\caption{Share of the change of the return predicted by the world model's gradient that appears when the same
perturbation is applied in the simulator (median over start states), and the same for the model's own finite
difference. Full-library 29-dof policies at iteration $25{,}000$; $5{,}456$ start states each, perturbation with
largest element $0.05$.}
\label{tab:trust}
\begin{tabular}{@{}lccccc@{}}
\toprule
$G$ / batch median & $0$--$1$ & $1$--$3$ & $3$--$10$ & $10$--$30$ & $>30$ \\
\midrule
simulator & $0.95$ & $0.89$ & $0.45$ & $0.07$ & $0.00$ \\
model, finite difference & $0.99$ & $0.99$ & $0.87$ & $0.22$ & $0.03$ \\
simulator, sign agrees & $1.00$ & $0.98$ & $0.88$ & $0.72$ & $0.60$ \\
\midrule
simulator, baseline's policy & $0.96$ & $0.84$ & $0.45$ & $0.08$ & $0.01$ \\
\bottomrule
\end{tabular}
\end{table}

\subsection{What the tests say about the update}

\emph{Trust is a matter of the norm, not of the state.} One might expect the model to be wrong where the motion is
hard. Within the band of $3$ to $10$ times the median, ranking states by the critic's value or by how far the policy
falls short of the reference from them, the third where the policy tracks worst has a trust of $0.41$ to $0.46$ and
the third where it tracks best $0.39$ to $0.41$ ($0.38$ to $0.42$ and $0.43$ to $0.49$ for the baseline's policy). The
large gradients are common on hard states, but a gradient of a given norm is about as reliable there as elsewhere.

\emph{The rule.} Both tests give a rollout of norm $G$ a worth near $\min(G, k \cdot \med G)$ with $k$ of a few
units. An update that weights every rollout by its full gradient spends most of its step on the part neither test
backs. An update that normalizes every rollout to the same norm discards the difference between a gradient of
$0.3$ medians and one of $3$, which both tests say is real, and on a subset of the library it leaves more than ten
times as many clips persistently failing in training as the cap (Appendix~\ref{app:subset}). The cap sits between
the two.

\emph{The level.} The tests do not single out a value of $k$. A lower cap removes more of the worth and leaves
less noise. Capping at $10$ removes $15\%$ of $\sum_b G_b$ but $1\%$ ($-2$ to $2\%$) of the summed worth; capping at
$3$ removes $25\%$ and $6\%$ ($3$ to $8\%$), and leaves $1.9$ times less energy $\E[\min(G, k\,\med G)^2]$, which
above a few medians is mostly noise. We use $k = 10$, the level of these two that removes least of the worth.

\section{Method}
\label{sec:method}

\begin{figure*}[t]
\centering
\includegraphics[width=\textwidth]{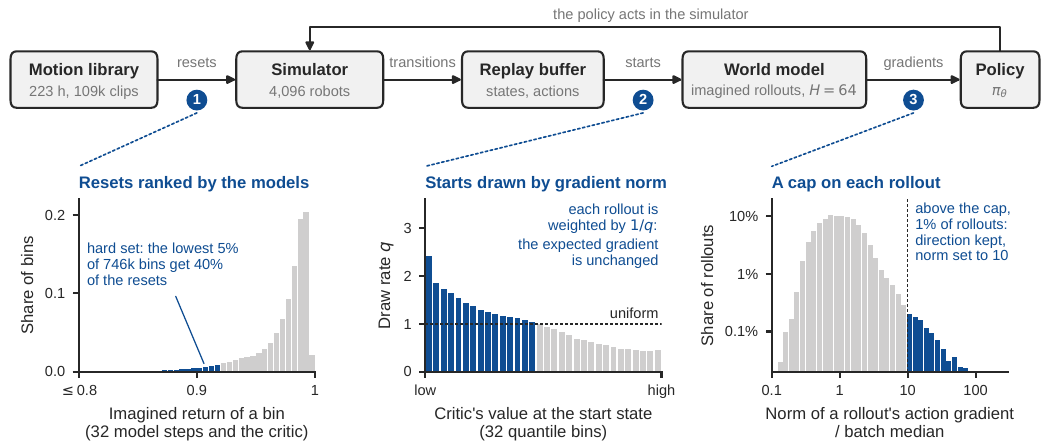}
\caption{The training loop and the three places the method changes it, each shown on the quantity it acts on.
\textbf{1}: every bin of the training library is ranked by its imagined return, and the lowest $5\%$ receive $40\%$
of the resets (plotted from a full-library run at iteration $25{,}000$). \textbf{2}: imagination starts are drawn at a rate
set by a table of the gradient norm per quantile of the critic's value, here $2.4$ times uniform for the
lowest-value states and $0.43$ times for the highest, and each rollout is weighted by the inverse (the same run's
table). \textbf{3}: the norm of each rollout's action gradient is capped at $10$ times the batch median, which
binds on $1\%$ of rollouts (the same run).}
\label{fig:overview}
\end{figure*}

The method changes three things in \base{}: the weight of each rollout in the policy gradient, the distribution of
the states that imagination starts from, and the distribution of the frames that episodes are reset to. A fourth,
an average of the policy's weights, is used for evaluation only. Fig.~\ref{fig:overview} gives an overview, and
Appendix~\ref{app:parts} shows each part in detail.

\subsection{A cap on each rollout's gradient}

With $G_b$ the gradient norm of rollout $b$ (Sec.~\ref{sec:background}) and $\med G$ the median over the batch,
each rollout receives the weight
\begin{equation}
w_b \;=\; \min\!\Big(1,\; \frac{k \,\med G}{G_b}\Big), \qquad k = 10,
\label{eq:cap}
\end{equation}
and the policy ascends $\frac{1}{B}\sum_b w_b \nabla_\theta J_b$ with the weights held constant. This is per-sample
clipping at a quantile of the norms~\cite{abadi2016dpsgd,andrew2021adaptive}, applied to a rollout's action gradient. A rollout below
the cap contributes its gradient unchanged; one above it contributes its direction at the norm $k \,\med G$. The
cap is relative, so it needs no tuning to the scale of the reward or of the model, and it follows the gradient's
scale as training proceeds.

\emph{Computing it.} The weights depend on the gradients they scale, so one backward pass does not suffice. We add
a zero tensor to the action of every step, backpropagate $\sum_b J_b$ to these tensors only, which gives all
$g_{b,t}$ and hence $G_b$ at the cost of one pass through the model, and then evaluate the policy once more at the
stored observations and backpropagate $\sum_{b,t} w_b\, g_{b,t}^\top \pi_\theta(o_{b,t})$ into its parameters. By
Eq.~\eqref{eq:action-gradient} this is exactly the weighted gradient. The second pass touches the policy alone.

\emph{Why per rollout.} A cap on the norm of the whole batch gradient, as in standard gradient
clipping~\cite{pascanu2013difficulty}, rescales the step and leaves its direction to the largest rollouts. A cap on each
step's action gradient inside the backward pass, as in concurrent work~\cite{lu2026nablamimic}, changes the
direction of a rollout's gradient. The tests of Sec.~\ref{sec:analysis} are statements about a rollout as a whole,
and the rule acts on that unit: about $60\%$ of the variance of $\log\lVert g_{b,t}\rVert$ lies between rollouts,
so the norm of a step's action gradient is mostly set by the rollout it belongs to.
Measured offline against a large-batch reference, a per-step cap gives a gradient of similar quality
(Appendix~\ref{app:analysis}); we did not train with it.

\emph{Horizon, batch and learning rate.} With the cap we use $H = 64$ imagined steps and $B = 1{,}024$ rollouts in
place of $32$ and $4{,}096$. A longer horizon gives the policy more of the reference to plan over and a later
bootstrap, and is affordable once its heavier gradient tail is cut; fewer rollouts keep the number of imagined
steps per iteration the same. We also double the baseline's policy learning rate, a setting carried over from
early short runs in which it helped, and not tuned for the cap.

\subsection{Starts drawn by gradient norm}

The cap takes weight from the states whose rollouts have large gradients, and those are not a random subset: the
gradient norm is largely a function of the start state (rank correlation $0.95$ between two rollouts of a start).
Instead of cutting a large gradient, we can sample its state more often and weight each of its rollouts down, as in
importance sampling of training examples by gradient norm~\cite{zhao2015stochastic,katharopoulos2018not}. Let
$q(s)$ be the rate at which a replay state is drawn relative to uniform. Weighting each rollout by $1/q$ leaves the
batch gradient's expectation unchanged for any $q > 0$, and if $q$ is proportional to the gradient norm the
weighted contributions all have the same norm, so a cap finds nothing to remove. We call this the
\emph{start draw}.

We approximate this with a small table. Each iteration takes $8B$ candidate states from the replay buffer and
evaluates the critic on each. The candidates are divided into $32$ quantiles of the critic's value, and a table
holds for each quantile a running mean, over about $200$ updates, of the gradient norms of the rollouts started in it
(norms above $30$ medians count as $30$). The $B$ starts are drawn with $q$ proportional to the table's entry for
their quantile, kept within a factor of $30$ of the median quantile's. Every rollout is weighted by $1/q$ in the
policy gradient, before the cap of Eq.~\eqref{eq:cap} is applied to the weighted norms, and by the same $1/q$ in the
critic's loss. The
critic's value predicts the gradient norm because both are functions of how well the policy tracks from a state:
the table explains $39\%$ of the variance of $\log G$ at the method's policy and $49\%$ at the baseline's. With the
policy held fixed, the draw lowers what a cap removes of the worth from $1\%$ to none at $k = 10$ and from $6\%$ to
$3\%$ at $k = 3$; at the baseline's policy, whose tail is heavier, from $7\%$ to $3\%$ and from $22\%$ to $11\%$
(Appendix~\ref{app:analysis}).

\subsection{Model-ranked resets}
\label{sec:resets}

The baseline's reset distribution learns which bins are hard from episodes started in them. On a large library
that is slow: a bin's score changes only when it is visited, and at $4{,}096$ environments the $746{,}000$ bins of
our training library are swept once in about $25{,}000$ iterations. The world model and the critic can score every
bin without visiting it. Every $2{,}000$ iterations we start an imagined rollout at the reference state of each
bin, follow the policy for $32$ steps and bootstrap with the critic, as in Eq.~\eqref{eq:objective}; the lowest
$5\%$ of bins by this imagined return form the \emph{hard set}. Resets keep the baseline's mix and go to a uniformly drawn bin
with probability $0.5$, to a bin of the hard set with probability $0.4$, and to a bin of earlier hard sets with
probability $0.1$. No episode is started in the last $50$ frames of a clip, where too little of the motion is left
for a return to say how well it was tracked. The ranking is cheap:
scoring all bins of our training library takes about $30$\,s, and the twelve rankings of a run add $3\%$
to its training time.

On the test motions the lowest imagined bin of a clip predicts whether the policy fails on it with an area under the
ROC curve of $0.97$ to $0.98$, against $0.96$ to $0.97$ for the critic's value alone, which is the signal of
critic-based curricula~\cite{tzannetos2023procurl,eimer2021space} (each of the six main runs, mean over its last six snapshots).
On the hardest tenth of unseen clips in earlier runs on a subset of the library the margin was larger, $0.91$ to $0.92$ against $0.74$ to $0.77$.

\subsection{An average of the policy for evaluation}

The policy that acts in training is noisy from update to update: at the last checkpoint an exponential average of its
weights with a time constant of $1{,}000$ iterations is $0.014$ above it in test return for the baseline and $0.010$ for
the method, and across the method's three seeds the acting policy's return ranges over $0.005$ and the average's over $0.0001$. We keep such an average from iteration $5{,}000$ and evaluate it. It never
acts in training and is applied to the baseline as well, so no difference we report comes from it.

\section{Experiments}
\label{sec:experiments}

\subsection{Setup}

\emph{Data and split.} We use BONES-SEED~\cite{bonesseed}, a library of $142{,}220$ motion-capture clips, without
the clips on stairs, ladders and chairs,
retargeted to the Unitree G1 with 29 and with 23 actuated joints by a whole-trajectory solver whose validator
rejects clips with penetration, unsupported frames or discontinuities. For the 29-dof robot $121{,}061$ clips pass.
We split them by \emph{motion}: all recordings of a motion by all actors, and their mirrored versions, go to the
same side. Of $3{,}578$ motions, the motions of each of the library's eight packages are ordered by their number of clips,
and every tenth, from a random offset, is held out. The training library has $109{,}233$ clips ($222.7$\,h) and the test library $11{,}828$ clips of $353$ motions
($21.3$\,h) that no actor performed in training. Both robots share the split.

\emph{Protocol.} The main runs train for $25{,}000$ iterations with $4{,}096$ simulated robots and is evaluated on
every clip of the test library, played from its first frame to its last by the deterministic policy without
exploration noise. We report the normalized return averaged over clips; its mean over the lowest $10\%$ and $1\%$
of a run's clips; and the number of clips that end in a fall in at least three of the six snapshots taken every $1{,}000$
iterations from $20{,}000$ to $25{,}000$, which we call \emph{persistently failing}. All returns are of the averaged
policy and are means over those six snapshots unless a single iteration is named. The lowest $10\%$ and $1\%$ in the
tables are of each run's own clips; a gain on the worst clips in the text compares two methods on the same clips,
the worst by the mean of the two. Confidence intervals on
differences between methods are $95\%$ bootstrap intervals that resample motions, not clips, since the clips of a
motion are not independent.

\emph{Baseline and hardware.} The baseline is \base{} with its published settings, to which we add only the
policy average of Sec.~\ref{sec:method}, for evaluation. It is our reimplementation: the original simulates with MuJoCo Warp~\cite{mujocowarp}, which we run
through Newton~\cite{newton}. We wrote it from the paper and then matched the inputs of the policy and the critic to the
code the authors released~\cite{supermimiccode}: the policy does not see the reference root's position relative to the robot, and the
critic does not see the previous action. All results reported here are of runs with these inputs. Both methods share the environment, networks, world-model updates and critic, so every difference we report is
between two updates in one code base. Each run uses one RTX~6000~Ada.

\subsection{Main result}

\begin{table}[t]
\centering
\caption{Test motions never performed in training, Unitree G1 with 29 dof, three seeds per method. Returns are of
the averaged policy over iterations $20{,}000$ to $25{,}000$ unless an iteration is named.}
\label{tab:main}
\begin{tabular}{@{}lcc@{}}
\toprule
 & \base{} & Ours \\
\midrule
Normalized return & $0.9475$ & $\mathbf{0.9603}$ \\
\quad per seed & $.9472, .9484, .9468$ & $.9602, .9602, .9603$ \\
Lowest $10\%$ of clips & $0.795$ & $\mathbf{0.869}$ \\
Lowest $1\%$ of clips & $0.640$ & $\mathbf{0.753}$ \\
Persistently failing clips & $17, 19, 14$ & $\mathbf{7, 3, 4}$ \\
\midrule
Return at iteration $25{,}000$ & $0.9486$ & $\mathbf{0.9613}$ \\
\quad on training motions & $0.9619$ & $\mathbf{0.9695}$ \\
\quad raw policy & $0.9344$ & $\mathbf{0.9515}$ \\
\midrule
Iterations to $0.9486$ & $25{,}000$ & $\mathbf{10{,}800}$--$\mathbf{11{,}000}$ \\
Hours to $0.9486$ & $2.5$--$2.6$ & $\mathbf{1.2}$--$\mathbf{1.3}$ \\
Seconds per iteration & $\mathbf{0.35}$--$\mathbf{0.37}$ & $0.39$--$0.42$ \\
\bottomrule
\end{tabular}
\end{table}

\begin{figure*}[t]
\centering
\includegraphics[width=\textwidth]{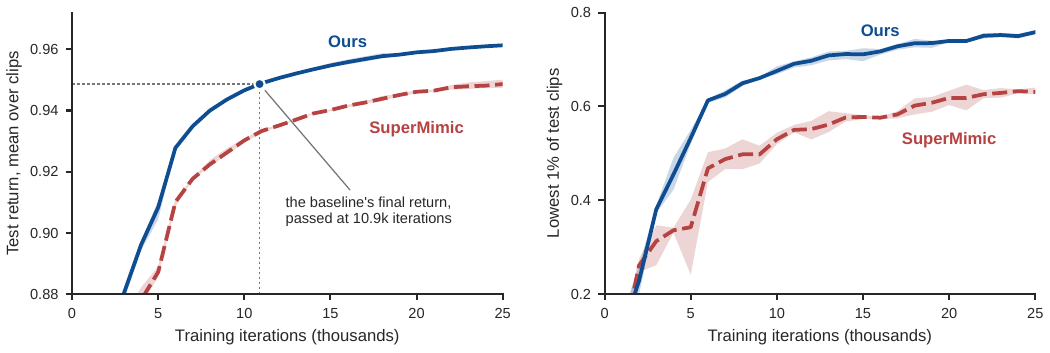}
\caption{Test return (left) and the mean return of the lowest $1\%$ of test clips (right) during training, for
\base{} (red, dashed) and our method (blue). Lines are means over three seeds and bands their range. The raw policy
is evaluated up to iteration $4{,}000$ and the averaged one from $5{,}000$.}
\label{fig:curves}
\end{figure*}

Table~\ref{tab:main} and Fig.~\ref{fig:curves} give the main comparison. The method raises the test return from
$0.9475$ to $0.9603$, a difference of $+0.0128$ ($95\%$ interval $+0.0071$ to $+0.0207$), and is higher on $89\%$
of clips and $90\%$ of motions. The distance to perfect tracking falls by $24\%$. Seeds of one method differ by
at most $0.0016$, an eighth of the difference between the methods.

\begin{figure}[t]
\centering
\includegraphics[width=\columnwidth]{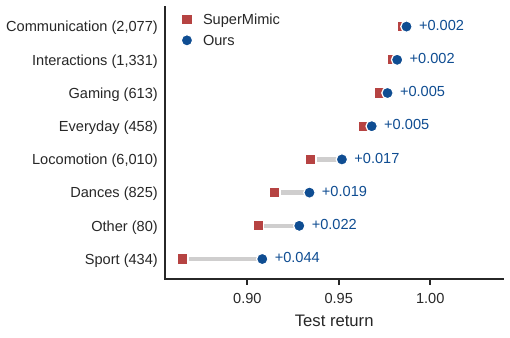}
\caption{Test return by package of the library, with the number of test clips in brackets, for \base{} (red
squares) and our method (blue circles), and the gain.}
\label{fig:packages}
\end{figure}

\emph{Hard motions.} The gain is largest where tracking is worst: $+0.078$ ($+0.051$ to $+0.097$) on the worst
$10\%$ of clips and $+0.006$ on the rest. The lowest $1\%$ of clips rises from $0.64$ to $0.75$, and the
persistently failing clips fall from $14$--$19$ to $3$--$7$. By package of the library the gain is $+0.044$ on sport,
$+0.019$ on dances and $+0.017$ on locomotion, and $+0.002$ to $+0.005$ on communication, interactions, gaming and everyday
motions, which the baseline already tracks at $0.96$ to $0.99$ (Fig.~\ref{fig:packages}). Longer clips gain more:
$+0.010$ under $4$\,s and $+0.021$ over $15$\,s.

\emph{Speed.} The method passes the baseline's test return at iteration $25{,}000$, $0.9486$, at iteration
$10{,}773$, $10{,}886$ and $11{,}006$
on its three seeds, after $1.2$ to $1.3$ hours; the baseline's $25{,}000$ iterations take $2.5$ to $2.6$ hours.
An iteration of the method costs $12\%$ more on average, and the cost is the cap's. By phase (medians over training), the
policy update, which includes the second pass of the cap, takes $0.21$ to $0.22$\,s against $0.16$\,s, and
collecting data and updating the world model and critic take $0.17$\,s against $0.19$\,s. A run with
$H = 64$, $B = 1{,}024$ and no cap updates its policy in $0.15$ to $0.16$\,s
and iterates as fast as the baseline, so the longer horizon on fewer rollouts costs no time. The reset rankings add
$3\%$ to a run.
After as many hours as the baseline trains, the method is at $0.961$.

\emph{Generalization.} At iteration $25{,}000$ the gap between the return on training motions and on test motions
is $0.008$ for the method
and $0.013$ for the baseline, so the gain does not come from fitting the training library more closely.

\emph{Failures.}\label{sec:failures} Two test clips fail persistently for both methods on every seed, a handstand
and its mirror image, at returns of $0.23$ to $0.25$. The cause is the data, not either update. The retargeter records the joint
torques its inverse dynamics needs to play each reference; divided by the G1's actuator limits, the handstand needs
up to $5.8$ times a limit and exceeds one on $74\%$ of its frames,
at the wrists, whose motors give $5$\,Nm, as the hands carry the body. Every clip that fails persistently
for our method on every seed is among the $1.9\%$ of test clips whose reference needs more than three times a limit.
Exceeding a limit does not by itself make a clip fail: $94\%$ of the clips above three times
are tracked by both methods on every seed. Burpees need $2.7$ to $3.8$ times a limit at the wrists as well, and here the
methods differ: six burpee clips fail persistently for the baseline on every seed, at returns of $0.10$ to $0.37$, and
for the method on one seed of one clip, at $0.55$ to $0.71$. Our filter removed motions on stairs, ladders and chairs and kept these; a filter on
the reference's torque would remove them. Four clips of two sitting motions with bent legs fail for the baseline on at
least two seeds and on none for the method, at a return of $0.84$ against $0.59$.

\begin{figure*}[t]
\centering
\includegraphics[width=\textwidth]{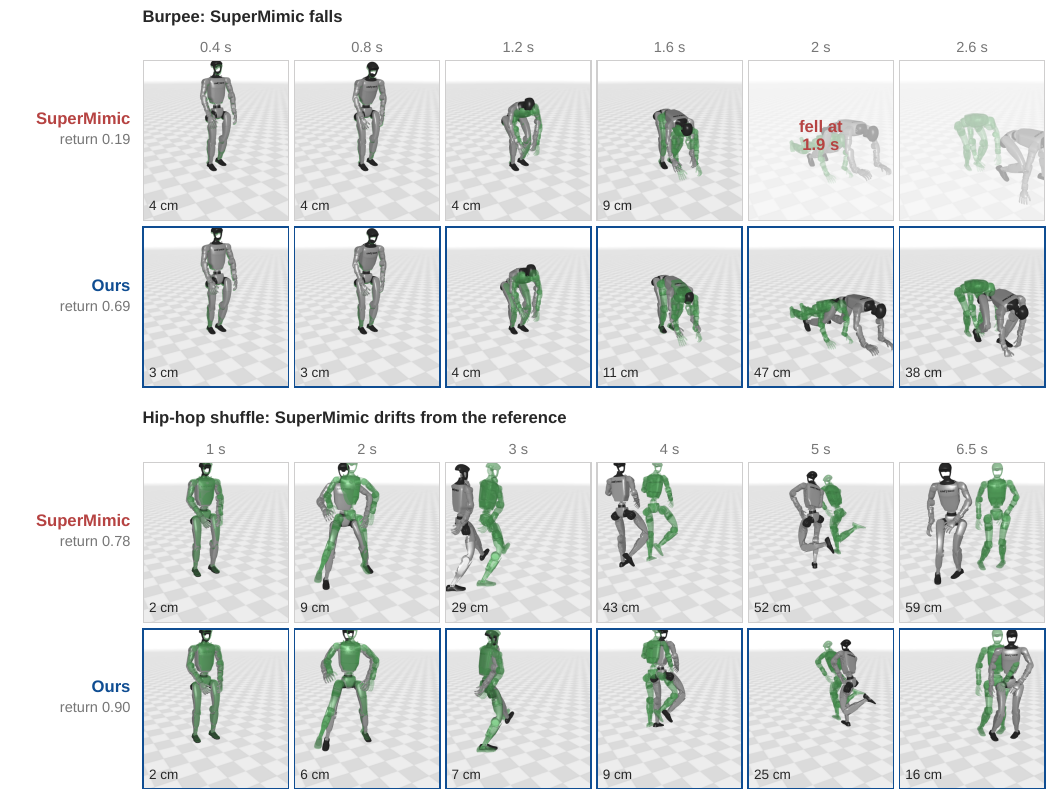}
\caption{Two test motions, never performed in training, played by \base{} (top row of each pair) and our method
(bottom row), with the reference in translucent green; the numbers in the frames are the mean position error of the
tracked bodies. Top: one of the four test clips that the baseline fails on all three seeds at iteration
$25{,}000$ and our method on none, all of them burpees. Bottom: the clip with the largest difference on the first seed among those of the
hip-hop shuffle, one of the seven motions with the largest gain, on which both complete the motion. The policies are
the averaged ones at iteration $25{,}000$ of the first seed. Frames are rendered from rollouts in stock MuJoCo, not
the training simulator; outcomes and returns (left) agree with the benchmark's to within $0.018$. After the
baseline's fall the rollout goes on for one second (faded frames), outside the episode.}
\label{fig:rollouts}
\end{figure*}

Fig.~\ref{fig:rollouts} shows two test clips played by the first seed of both methods: one on which the baseline
falls, and one on which both complete the motion and the baseline tracks it less closely.

\subsection{What each part does}
\label{sec:ablations}

\begin{table}[t]
\centering
\caption{Ablation on the 29-dof test set at iteration $15{,}000$ (averaged policy over iterations $10{,}000$ to
$15{,}000$), three seeds per row. Returns are means over seeds; tails are of each run's own clips, means over
seeds; failing is the range over seeds of the clips failing in three or more of the six snapshots. Each row names
the parts it uses, and where resets are not named they are the baseline's. The upper rows use the baseline's
horizon, batch and learning rate unless stated, the lower ones $H = 64$, $B = 1{,}024$ and the doubled learning
rate.}
\label{tab:ablation}
\setlength{\tabcolsep}{3.5pt}
\begin{tabular}{@{}lccccc@{}}
\toprule
 & return & low $10\%$ & low $1\%$ & failing & $-$ base \\
\midrule
\multicolumn{6}{@{}l}{\emph{$H = 32$, $B = 4{,}096$}} \\
\base{} & $.9358$ & $.749$ & $.575$ & $21$--$28$ & \\
\quad learning rate $\times 2$ & $.9380$ & $.755$ & $.574$ & $18$--$23$ & $+.0022$ \\
\quad model-ranked resets & $.9396$ & $.813$ & $.670$ & $12$--$16$ & $+.0038$ \\
\midrule
\multicolumn{6}{@{}l}{\emph{$H = 64$, $B = 1{,}024$, learning rate $\times 2$}} \\
\quad resets, no cap & $.9204$ & $.759$ & $.647$ & $\mathbf{3}$--$\mathbf{12}$ & $-.0153$ \\
\quad cap, start draw & $.9465$ & $.786$ & $.625$ & $18$--$24$ & $+.0107$ \\
\quad cap, resets & $.9504$ & $.838$ & $.700$ & $12$--$14$ & $+.0146$ \\
Ours (cap, draw, resets) & $\mathbf{.9511}$ & $\mathbf{.840}$ & $\mathbf{.706}$ & $5$--$16$ & $\mathbf{+.0153}$ \\
\bottomrule
\end{tabular}
\end{table}

Table~\ref{tab:ablation} removes the parts one at a time, at $15{,}000$ iterations, and adds two controls for the
settings the method changes besides its three parts: the doubled learning rate alone, and the method's horizon,
batch and learning rate without the cap.

\emph{The cap gives the mean.} With the baseline's own resets, the cap and the start draw raise the return by
$+0.0107$ ($+0.0084$ to $+0.0135$), on $98\%$ of clips. They do not reduce the persistently failing clips ($18$ to
$24$ against $21$ to $28$).

\emph{Not the learning rate or the horizon.} Doubling the baseline's learning rate raises its return by $+0.0022$
($+0.0010$ to $+0.0035$), on $71\%$ of clips, and leaves the failing clips about where they were ($18$ to $23$); our
method is $+0.0130$ ($+0.0069$ to $+0.0209$) above it. The method's horizon, batch and learning rate with the
plain gradient, with the model-ranked resets, are $0.0153$ below the baseline and $0.0192$ ($0.0148$ to $0.0243$)
below the resets alone. Adding the cap to that run, its only difference from the row with the cap and resets,
raises the return by $+0.0300$ ($+0.0246$ to $+0.0366$), on all but $0.4\%$ of clips. The longer horizon pays off
only with the cap. Without it, that run leaves the fewest clips persistently failing ($3$ to $12$), but its worst $1\%$
are tracked worse than with the cap ($0.647$ against $0.700$). We have not found why the run with the lowest return
leaves the fewest clips failing. Its low return may partly come from the doubled learning rate, which was chosen
for a normalized gradient in early runs and may not suit the plain one.

\emph{The model-ranked resets give the tail.} Adding the model-ranked resets to the cap raises the worst $10\%$ of clips
by $+0.055$ ($+0.037$ to $+0.069$) and the lowest $1\%$ from $0.63$ to $0.71$, and cuts the failing clips from $18$--$24$
to $5$--$16$, at a cost of $0.001$ on the other $90\%$ of clips. Alone, on the baseline's update, the resets raise
the worst $10\%$ by $+0.067$ and lower the rest by $0.003$; their effect on the mean, $+0.0038$ ($-0.0016$ to
$+0.0112$), is not distinguishable from zero. Resetting to hard motions trades a little on the easy ones, and only
with the cap does the policy turn the extra data into a higher mean.

\emph{The size of the hard share.} The mix of resets ($0.5$ uniform, $0.4$ hard, $0.1$ earlier hard sets) is
\base{}'s; only the choice of the hard bins differs. In earlier runs with the baseline's update on the whole 23-dof library, made before we matched the inputs to the released code, one seed each and judged on the
training clips, halving the hard share to $0.2$ raised the mean by $0.0011$ but left four clips failing in four or
more of six snapshots, against none; keeping the share and widening the hard set from the lowest $2\%$ of bins to
$5\%$, the setting we use, raised it by $0.0018$ with none failing. The share trades mean for tail, and spreading
it over more bins raised the mean without giving up the tail.

\emph{The start draw adds little at this scale.} Without the draw the return is $0.9504$ against $0.9511$, a
difference of $+0.0007$ ($+0.0003$ to $+0.0011$), and $+0.0028$ ($+0.0009$ to $+0.0048$) on the worst $10\%$ of
clips. The lowest $1\%$ is $0.700$ against $0.706$ and the failing clips $12$ to $14$ against $5$ to $16$, and the
run without the draw passes the baseline's final return almost as early ($11{,}149$ to $11{,}395$ iterations against
$10{,}773$ to $11{,}006$). In training the draw does what it was built to do, drawing starts from the most-drawn
quantile of the critic's value $2.4$ to $2.7$ times as often as from the median one, and its work is done early
in training. At the method's policy of iteration $4{,}000$ a cap at $10$ removes $3.6\%$ ($2.5$ to $4.8\%$) of the
worth, and drawing the starts lowers this to $1.3\%$ ($0.6$ to $2.2\%$); at iteration $25{,}000$ the cap removes
$0.9\%$, and what the draw changes is within the noise (Appendix~\ref{app:analysis}). The tail is
no heavier early on, but its gradients are worth more, $4.5$ medians from $5$ to $30$ times the median against
$2.6$, so the cap has more to remove. This fits the slightly later crossing of the run without the draw. Its gain is larger where the resets are the baseline's: on a subset of the
library, in earlier runs, where the cap alone left $43$ clips persistently failing, the draw reduced them to $2$
(Appendix~\ref{app:subset}). We read this as two ways of directing the update to hard states, and the model-ranked
resets already do most of what the draw does.

\emph{The cap level.} On a $15\%$ subset of the 23-dof library, in earlier runs in which the policy also saw the reference root's position, we trained caps at $3$, $5$, $7$, $10$ and $30$
times the median (Appendix~\ref{app:subset}). With the start draw and the resets, caps from $3$ to $7$ are $0.0003$
to $0.0006$ above $10$ in mean return but leave up to four clips persistently failing, against one or two at $10$.
With the resets and without the draw, a cap at $30$ is $0.0014$ below a cap at $10$. A lower cap leaves less noise
in the update and removes more of the worth, and from $10$ down to $3$ the two effects come to $+0.0011$ and
$-0.0007$ in return. The worth a cap removes lies on the states where gradients are large, which are hard states,
so the mean is flat in $k$ and the tail is not. Normalizing every rollout to the median, the limit of a low cap,
leaves $36$ to $42$ clips persistently failing on the subset against $2$ to $3$ for the cap at $10$.

\emph{What did not help.} On the subset we also tried a cap on the mean gradient of each start over two rollouts
($-0.0022$), returning the removed part of the gradient through a slow average ($-0.0004$ to $-0.0041$), a longer
momentum on the uncapped gradient, and an adaptive rule that sets the weights each iteration, between the cap and
normalization, to reach a target effective sample size. None improved on the cap. In an earlier run on the subset, halving the world-model updates
per iteration saved $13\%$ of an iteration and cost $23\%$ more iterations.

\subsection{A second robot}

On the G1 with $23$ actuated joints, trained on its own retargeting of the same library ($109{,}708$ training clips,
$11{,}859$ test clips of the same $353$ motions) with one seed per method, the method reaches $0.9678$ against
$0.9599$, a difference of $+0.0079$ ($+0.0050$ to $+0.0119$); the lowest $1\%$ of clips is $0.77$ against $0.62$;
the persistently failing clips are $1$ against $13$;
and the method passes the baseline's final return at iteration $10{,}188$, after $1.13$ hours of the baseline's
$2.27$.

\subsection{A second motion set}
\label{sec:lafan}

\begin{table}[t]
\centering
\caption{LAFAN1, Unitree G1 with 29 dof, trained and evaluated on the same $40$ clips, three seeds per method.
Returns are of the averaged policy over iterations $10{,}000$ to $15{,}000$ unless an iteration is named.}
\label{tab:lafan}
\setlength{\tabcolsep}{3pt}
\begin{tabular}{@{}lcc@{}}
\toprule
 & \base{} & Ours \\
\midrule
Normalized return & $0.8402$ & $\mathbf{0.8718}$ \\
\quad per seed & $.8410, .8425, .8371$ & $.8715, .8708, .8731$ \\
Return at iteration $15{,}000$ & $0.8494$ & $\mathbf{0.8768}$ \\
\midrule
Iterations to $0.8494$ & $13{,}400$--$14{,}900$ & $\mathbf{7{,}000}$--$\mathbf{7{,}500}$ \\
Hours to $0.8494$ & $1.31$--$1.46$ & $\mathbf{0.78}$--$\mathbf{0.84}$ \\
Seconds per iteration & $\mathbf{0.35}$ & $0.39$--$0.40$ \\
\bottomrule
\end{tabular}
\end{table}

LAFAN1~\cite{harvey2020lafan} is the motion set on which \base{} and the systems that build on
it~\cite{chen2026blinkmimic,dao2026instanttrack} report results. We use Unitree's retargeting to the 29-dof
G1~\cite{unitree2024lafan}: $40$ clips of $15$ motions, each performed by up to five actors, $2.45$\,h in all and
$1.7$ to $4.6$ minutes per clip. It is retargeted with kinematic constraints only. We resample it to $50$\,Hz and
raise every root by $5$\,mm, because its feet otherwise sit that much lower than in BONES-SEED. As in that line of work, the methods
train and are evaluated on the same clips, so this measures how well each fits a small library, not
generalization. Both train for $15{,}000$ iterations with the settings of the main experiment, three seeds each,
and each clip is played whole from its first frame.

Table~\ref{tab:lafan} gives the result. The method reaches $0.8718$ against $0.8402$, a difference of $+0.0316$
($+0.0229$ to $+0.0412$, resampling motions), and is higher on $37$ of the $40$ clips and on all $15$ motions. The
distance to perfect tracking falls by $20\%$. The six clips of falling and getting up gain $+0.071$
on average ($+0.061$ to $+0.074$), against $+0.025$ for the other clips. The method passes the baseline's final
return at $7{,}000$ to $7{,}500$ iterations, after $0.78$ to $0.84$ hours; the baseline's seeds reach it at
$13{,}400$ to $14{,}900$, after $1.31$ to $1.46$ hours. Both methods are still improving at $15{,}000$ iterations, so this compares
them at one training budget.


\section{Limitations}
\label{sec:limitations}

\emph{The cap level is chosen, not derived.} The two tests locate where a gradient stops being worth its norm, at a
few times the median, and training is flat in the cap from $3$ to $10$. They do not give a value, and we did not
tune $k$ on the full library.

\emph{The baseline's settings.} The baseline runs at its published settings apart from the learning rate, which we
tested: doubled, it raises the baseline's return by $0.0022$ at $15{,}000$ iterations, about a seventh of the
method's gain there, and leaves its failing clips about where they were (Table~\ref{tab:ablation}). We have not tuned its
other settings, and the main comparison at $25{,}000$ iterations is against the published ones.

\emph{An iteration costs more.} The cap's second pass through the policy makes an iteration $12\%$ slower on the GPU
we trained on; without the cap, the method's horizon and batch cost what the baseline's do. Folding the two passes
into one, or capturing the update in a GPU graph as recent systems do~\cite{choi2026instantmimic,chen2026blinkmimic},
may remove part of that cost; we have not tried either.

\emph{What the tests measure.} The simulator test perturbs actions along one direction per state, the model's gradient, at two scales,
and it compares the model with the simulator it was trained on, not with a robot. Both tests are of policies
at the end of training; at an early checkpoint, with a model that has seen less data, the simulator confirms less in the middle
bands (Appendix~\ref{app:analysis}).

\begin{figure}[t]
\centering
\includegraphics[width=\columnwidth]{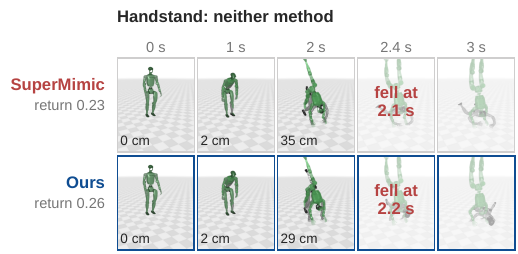}
\caption{A test motion neither method learns: going into a handstand, played as in Fig.~\ref{fig:rollouts}. Its
reference needs up to $5.8$ times the torque limit of the G1's wrist motors as the hands take the body's weight, and
both policies fall when the hands reach the ground.}
\label{fig:fail}
\end{figure}

\emph{Motions beyond the robot.} A handstand fails for both methods on every seed (Fig.~\ref{fig:fail}).
Its reference needs several times the torque of the G1's wrist motors (Sec.~\ref{sec:failures}), so the failure
comes from our data filter, which kept it, and not from either method. Filtering the library by the reference's
torque, which the retargeter already computes, would remove such clips. Since both methods train and are tested on the
same clips, this failure favors neither.

\section{Conclusion}

A learned world model gives an exact gradient of itself, and for most imagined rollouts of a humanoid tracker that
gradient is nearly the simulator's. For the few rollouts with the largest gradients it is not: neither a redraw of
the same state nor the simulator backs more than a few medians of it. Capping each rollout's gradient at that
level, drawing starts so that the cap has less to remove, and resetting episodes where the models predict failure
turns this observation into a tracker that follows unseen motions more closely, fails on fewer of them, and reaches
the baseline's quality in less than half the training time. The tests themselves need only a trained policy, its
world model and the simulator, and apply to any method that differentiates a learned model.

\section*{Acknowledgment}
Motion Data by Bones Studio (\url{https://bones.studio/}).

\bibliographystyle{IEEEtran}
\bibliography{refs}

\appendices
\section{Details of the analysis}
\label{app:analysis}

\emph{Policies.} The tests of Sec.~\ref{sec:analysis} use the first seed of the method and of the baseline on the
full 29-dof library at iteration $25{,}000$, the policy that acts in training, not its average. Start states are
replay states collected by the policy under test, for both policies under the method's reset distribution. Before the full-library runs the same
tests were made on policies trained on a stratified $15\%$ of the 23-dof library ($18{,}237$ clips), with the earlier inputs
under which the policy also saw the reference root's position: the method at
iterations $15{,}000$ and $4{,}000$, the cap without the start draw, and per-rollout normalization.

\emph{What a cap removes.} Table~\ref{tab:bias} gives the share of the summed norm and of the summed worth that a
cap removes at the method's policy, and the energy $\E[c^2]/\med^2$ of the capped contributions
$c = \min(G, k\,\med)$, which we call noise since above a few medians almost none of it repeats. At the baseline's policy, whose tail is heavier, a cap at $10$ removes $25\%$ of the norm and $7\%$ ($4$ to $9\%$)
of the worth, and a cap at $3$ removes $41\%$ and $22\%$ ($18$ to $25\%$); with the starts drawn, $3\%$ ($2$ to
$5\%$) and $11\%$ ($8$ to $14\%$) of the worth.

\emph{Early in training.} At the method's policy of iteration $4{,}000$, with start states from that policy's
replay buffer and the resets of iteration $25{,}000$'s hard set (no benchmark of the training clips exists at
$4{,}000$), the tail is no heavier than at $25{,}000$: $0.22\%$ of rollouts lie above $30$ times the median and hold
$9\%$ of the summed norm. It is worth more. A rollout between $5$ and $30$ times the median is worth $4.5$ ($4.0$ to
$5.0$) medians, against $2.6$ at iteration $25{,}000$, and $16\%$ of the gradient's energy repeats in a redraw,
against $2\%$. A cap at $10$ then removes $11\%$ of the norm and $3.6\%$ ($2.5$ to $4.8\%$) of the worth, $1.3\%$
($0.6$ to $2.2\%$) with the starts drawn; a cap at $3$ removes $23\%$ of the norm and $12\%$ ($11$ to $14\%$) of
the worth, $7\%$ ($6$ to $9\%$) with the starts drawn.

\begin{table}[h]
\centering
\caption{What a cap removes of the gradient's norm and of its worth (in brackets, $95\%$ interval), and the
noise energy it leaves. The method's full-library policy; $32{,}768$ start states, two rollouts each.}
\label{tab:bias}
\setlength{\tabcolsep}{4pt}
\begin{tabular}{@{}lccc@{}}
\toprule
rule & norm & worth & noise \\
\midrule
cap $3$ & $.25$ & $.06$ [$.03$, $.08$] & $2.1$ \\
cap $5$ & $.20$ & $.03$ [$.00$, $.05$] & $2.7$ \\
cap $10$ & $.15$ & $.01$ [$-.02$, $.02$] & $3.9$ \\
cap $30$ & $.09$ & $.00$ [$-.02$, $.01$] & $7.3$ \\
cap $3$, starts drawn & $.21$ & $.03$ [$.01$, $.05$] & $2.2$ \\
cap $5$, starts drawn & $.16$ & $.01$ [$-.01$, $.03$] & $2.8$ \\
cap $10$, starts drawn & $.12$ & $.00$ [$-.02$, $.01$] & $3.8$ \\
\bottomrule
\end{tabular}
\end{table}

\emph{The same tests on other policies.} On the subset's policies a rollout between $5$ and $30$ times the median
is worth $4.1$ ($3.0$ to $5.1$) medians for the method, $3.8$ ($2.7$ to $4.7$) for the cap without the draw and
$5.1$ ($4.4$ to $5.9$) for normalization. The median trust by band of $G$ ($0$--$1$, $1$--$3$, $3$--$10$,
$10$--$30$, above $30$ medians) is there $0.94$, $0.84$, $0.56$, $0.11$, $0.01$ for the method at iteration
$15{,}000$; $0.95$, $0.84$, $0.57$, $0.12$, $0.01$ for the cap without the draw; and $0.92$, $0.72$, $0.38$, $0.10$,
$0.00$ for the method at iteration $4{,}000$, when the model has seen less data: more data improves the middle
bands and not the tail. With a perturbation five times smaller, the method's full-library policy of Table~\ref{tab:trust}
gives $0.95$, $0.91$, $0.43$, $0.04$, $0.00$, and the baseline's $0.95$, $0.81$, $0.39$, $0.06$, $0.01$.

\emph{Per-step and per-rollout caps.} On six checkpoint and horizon combinations of earlier trackers we compared,
offline, a cap on each rollout with a cap on each step's action gradient, applied after the backward pass or inside
it, by the alignment of the resulting batch gradient with a reference from $4{,}096$ rollouts. The two caps are
within $0.02$ of each other in five of six cases; at an early checkpoint and a horizon of $128$ the per-rollout cap
is ahead ($0.76$ against $0.65$ to $0.71$). Between $58\%$ and $64\%$ of the variance of
$\log\lVert g_{b,t}\rVert$ lies between rollouts.

\section{The three parts in detail}
\label{app:parts}

\begin{figure*}[t]
\centering
\includegraphics[width=\textwidth]{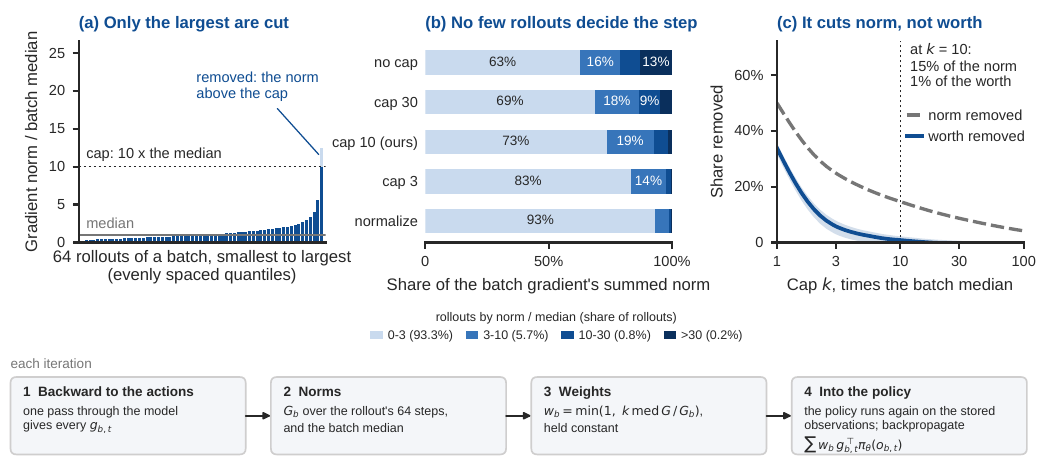}
\caption{The cap on each rollout's gradient ($32{,}768$ start states, two rollouts each). (a) The norms of a batch as
$64$ evenly spaced quantiles: a cap at $10$ times the median shortens only the largest, the $1\%$ of rollouts above
it. (b) The share of the batch gradient's summed norm held by the rollouts of each band of norm under each rule.
Without a cap, the $0.2\%$ of rollouts above $30$ times the median hold $13\%$; normalizing every rollout to the same
norm gives each band no more than its share of rollouts. (c) What a cap at $k$ removes of the summed norm (dashed)
and of the summed worth, the part that a second rollout from the same start repeats (solid, with its $95\%$
interval over starts). Below: the steps of one policy update.}
\label{fig:app-cap}
\end{figure*}

\begin{figure*}[t]
\centering
\includegraphics[width=\textwidth]{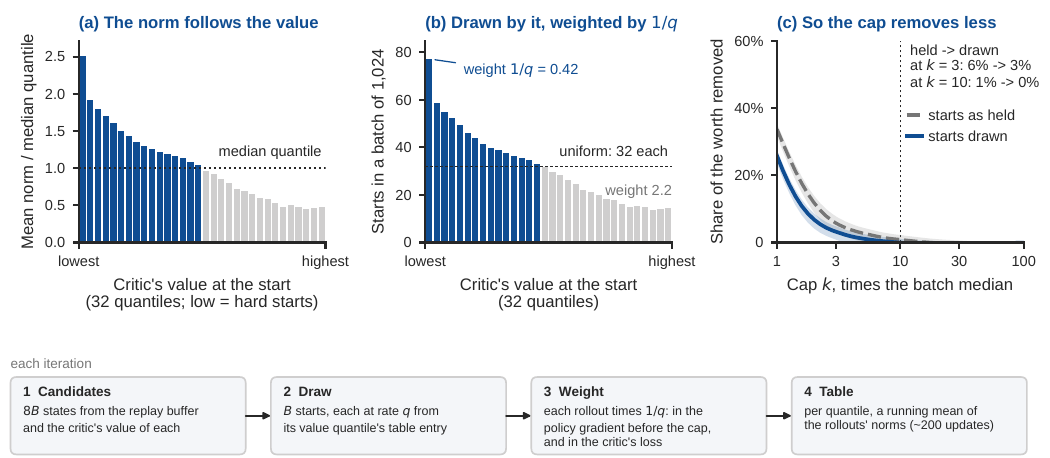}
\caption{The start draw. (a) The run's table: the mean gradient norm of the rollouts started in each quantile of the
critic's value, over that of the median quantile. Starts of low value, where the policy tracks worst, have up to
$2.5$ times the median norm. (b) The starts of a batch of $1{,}024$ drawn by that table, against $32$ per quantile
when drawn uniformly, and the weight $1/q$ each of their rollouts carries. (c) The share of the worth that a cap
removes with the starts as the replay buffer holds them (dashed) and drawn by a table fitted to the same rollouts
(solid), each with its $95\%$ interval, as in Table~\ref{tab:bias}. Below: the steps of one draw.}
\label{fig:app-draw}
\end{figure*}

\begin{figure*}[t]
\centering
\includegraphics[width=\textwidth]{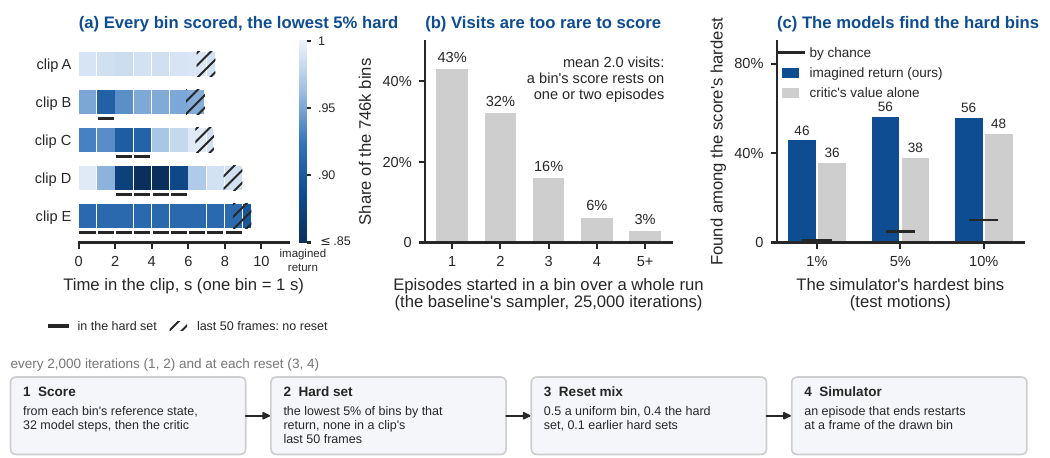}
\caption{The model-ranked resets. (a) Five training clips, each bin coloured by its imagined return. Underlined bins
are in the hard set, the lowest $5\%$ of the library's bins, and no reset starts in the hatched frames. (b) How many
episodes the baseline's sampler starts in a bin over its whole run of $25{,}000$ iterations: $75\%$ of the bins are
visited once or twice, so a score from visits rests on one or two episodes. (c) On the bins of the test motions,
the share of the simulator's hardest $1\%$, $5\%$ and $10\%$, by the return of a rollout started in the bin, that
each score places among its own hardest share; dashes mark chance. Below: the steps of a ranking and of a reset.}
\label{fig:app-resets}
\end{figure*}

Figs.~\ref{fig:app-cap}--\ref{fig:app-resets} show each part of the method on the first seed of its full-library 29-dof
run at iteration $25{,}000$, with the steps it adds to an iteration.

\section{Experiments on a subset of the library}
\label{app:subset}

Before the full-library runs, the parts of the method were developed on a stratified $15\%$ of the 23-dof library,
training for $15{,}000$ iterations and evaluating on the clips trained on (Table~\ref{tab:subset}). These runs predate the
inputs of Sec.~\ref{sec:experiments}: their policies also saw the reference root's position relative to the robot, and we
did not repeat them.

\begin{table}[h]
\centering
\caption{Subset of the 23-dof library ($18{,}237$ clips), averaged policy over iterations $10{,}000$ to $15{,}000$,
evaluated on the training clips. Two entries are two seeds.}
\label{tab:subset}
\setlength{\tabcolsep}{4pt}
\begin{tabular}{@{}lccc@{}}
\toprule
 & return & failing & low $1\%$ \\
\midrule
\base{} & $.9666$ & $4$ & $.736$ \\
model-ranked resets & $.9682$ & $0$ & $.828$ \\
normalize, resets & $.9739$, $.9741$ & $42$, $36$ & $.713$, $.745$ \\
cap $10$ & $.9734$ & $43$ & $.698$ \\
cap $10$, start draw & $.9747$ & $2$ & $.790$ \\
cap $3$, resets & $.9755$, $.9752$ & $21$, $36$ & $.786$, $.769$ \\
cap $10$, resets & $.9755$, $.9754$ & $2$, $3$ & $.841$, $.844$ \\
cap $30$, resets & $.9741$ & $0$ & $.844$ \\
cap $3$, draw, resets & $.9764$ & $4$ & $.842$ \\
cap $5$, draw, resets & $.9764$, $.9763$ & $0$, $4$ & $.856$, $.845$ \\
cap $7$, draw, resets & $.9762$ & $1$ & $.855$ \\
cap $10$, draw, resets (ours) & $.9759$, $.9758$ & $2$, $1$ & $.850$, $.854$ \\
\bottomrule
\end{tabular}
\end{table}

\emph{Why the cap level is flat.} Over the eleven runs of the seven cap rules in Table~\ref{tab:subset}, the mean
return is described to $10^{-4}$ by a plane in the noise energy $N$ a rule leaves and the share $R$ of the
worth it removes, both measured on the subset's policies: $0.9785 - 0.00033\,N - 0.0119\,R$. From a cap of $10$ to a cap of
$3$ on drawn starts the two terms are $+0.0011$ and $-0.0007$; without the draw they are $+0.0012$ and $-0.0014$.
On the hardest quarter of clips the coefficient of $R$ is ten times that on the other three quarters and the
coefficient of $N$ four times, which is why a lower cap leaves the mean unchanged and costs the tail.

\section{Settings}
\label{app:settings}

\begin{table}[h]
\centering
\caption{Settings. Everything not listed is \base{}'s.}
\label{tab:settings}
\setlength{\tabcolsep}{2pt}
\small
\begin{tabular}{@{}lcc@{}}
\toprule
 & \base{} & Ours \\
\midrule
Imagination horizon $H$ & $32$ & $64$ \\
Imagined rollouts $B$ & $4{,}096$ & $1{,}024$ \\
Policy learning rate & $2\times10^{-4}$ & $4\times10^{-4}$ \\
Cap $k$ on $G_b / \med G$ & -- & $10$ \\
Start candidates per rollout & $1$ & $8$ \\
Start-draw quantiles, memory & -- & $32$, $200$ updates \\
Start draw rate, range & -- & $[1/30, 30]$ \\
Reset mix (uniform, hard, past) & $0.5$, $0.4$, $0.1$ & $0.5$, $0.4$, $0.1$ \\
Hard bins by & visit score & lowest $5\%$, imagined \\
Ranking: horizon, interval & -- & $32$ steps, $2{,}000$ it. \\
Ranking: bins scored at a time & -- & $32{,}768$ \\
No reset in a clip's last & -- & $50$ frames \\
\midrule
Simulator & \multicolumn{2}{c}{Newton (MuJoCo Warp), $200$\,Hz} \\
Simulated robots, control & \multicolumn{2}{c}{$4{,}096$, $50$\,Hz} \\
World model & \multicolumn{2}{c}{MLP $3\times2{,}048$; $10$ batches per it.} \\
Policy encoder, decoder & \multicolumn{2}{c}{$2048$-$1024$-$512$-$256$ each} \\
Critic & \multicolumn{2}{c}{MLP $256$-$128$, $5$ epochs per it.} \\
Discount $\gamma$, $\lambda$ & \multicolumn{2}{c}{$0.99$, $0.95$} \\
Replay & \multicolumn{2}{c}{$1{,}000$ steps per robot} \\
Policy average (eval.) & \multicolumn{2}{c}{time const. $1{,}000$ it., from $5{,}000$} \\
\bottomrule
\end{tabular}
\end{table}

\end{document}